\documentclass[10pt,twocolumn,letterpaper]{article}

\usepackage{cvpr}
\usepackage{times}
\usepackage{epsfig}
\usepackage{graphicx}
\usepackage{amsmath}
\usepackage{amssymb}
\usepackage{subcaption}
\usepackage{geometry}
\usepackage[pagebackref=true,breaklinks=true,letterpaper=true,colorlinks,bookmarks=false]{hyperref}

\cvprfinalcopy 

\def\cvprPaperID{****} 

\ifcvprfinal\fi
\begin{document}

\title{Advanced Deep Learning Techniques for Analyzing Earnings Call Transcripts: Methodologies and Applications}
\author{Umair Zakir Abowath\\
Georgia Institute of Technology\\
\and
Evan Daykin\\
Georgia Institute of Technology\\
\and
Amssatou Diagne\\
Georgia Institute of Technology\\
\and
Jacob Faile\\
Georgia Institute of Technology\\
}

\maketitle

\begin{abstract}
   This is a comparative study of three deep learning models—BERT, FinBERT, and ULMFiT—in sentiment analysis of earnings call transcripts. The goal is to find out how leveraging Natural Language Processing (NLP) to extract sentiments conveyed in large financial transcripts can help in better-informed investment decisions and risk management strategies. We explore the advantages and disadvantages each of the deep learning model offers in the context of our task. We will also assess the different requirements of each model in regard to the data pre-processing as well as the optimization of their computational resources. Through rigorous experimentation and analysis, we evaluate and perform a comparative study of the models using various performance metrics such as accuracy, precision, recall, and F1-score. Finally, we will discuss what possible improvements can help make these models more suitable for sentiment analysis in financial contexts.
\end{abstract}
\section{Introduction/Background/Motivation}
\subsection{Objective}
 In this paper, we attempt to take one piece of the universe of public market data (equity earnings call transcripts) and seek to correlate the overall sentiment therein with the actual future performance of the company. We seek to test whether or not 'peripheral' indicators of performance, in this case, the neural network derived sentiment of earnings transcripts, can reliably predict future results in excess of the broader market.

\subsection{Current Practice}
\textit{Is it really priced in?} The weak-form Efficient-Markets hypothesis, at this point more an aphorism than anything rigorous or scientific \footnote{Author's opinion}, posits that any and all publicly available information is factored into the price of any given security on the open market. The theory goes that in the long run, there are no arbitrage opportunities in such a market, and there are no 'bargains'\cite{fama1970efficient}. Due to the empirical and behavioral-economic doubt cast on this hypothesis\cite{Siegel2010BlackSO}, we think it is possible that there still exists a market inefficiency that can be exploited.

If a natural-language neural network is able to reliably beat the major indices after adjusting for risk, it presents an opportunity to realize a net profit, with a competitive advantage over other investors in the market.

\subsection{Data Collection}
To turn this problem into a straight-forward supervised learning task, we needed two things: The transcripts themselves, and samples of the underlying market performance. 

For a good sampling of transcripts, we wanted the most recent four quarters' earnings reports for the constituents of the S\&P 500 \cite{sp500}, an index seeking to track the overall performance of the large-cap U.S. equities. Since there is not a simple way to export earnings transcripts \textit{en masse} from Bloomberg and Refinitiv (and is against their Terms of Service), we created a Python application controlling a ChromeDriver \cite{chromedriver} instance. The script, using a standard chrome browser, navigates to the HTML browser endpoints on SeekingAlpha \cite{seekingalpha} for each earnings call, saves the HTML, strips view elements, and returns the call in plain text format.

With the transcripts in hand, we then turn to pulling the market data for each company. Encoded in the HTML version of the transcript is the date of the earnings report. Using the AlphaVantage financial markets API \cite{alphavantage}, we retrieved the nearest share price quote and S\&P index benchmark to 90 days before earnings, 2 days before earnings, 2 days after earnings, and 90 days after earnings.

Using the previously-gathered data, we applied a multi-step transformation to the transcript data. Since all companies are going to say nice things about themselves, we cut off everything before the Q\&A segment of the calls, so a sentiment analysis can focus just on whether or not these officers are being "grilled" by their investors. Next, we converted the text to lower-case and removed punctuation. After this, stop-words (\eg like, as, and) from the Python Natural Language Toolkit (\texttt{nltk}) english stop-word corpus \cite{nltk} were removed from the data. Finally, we used the nltk \textit{id.} implementation of the Porter stemming algorithm \cite{porter1980algorithm} to remove suffixes from the text. For example, the words 'connected', 'connecting' and 'connects' all become 'connect'.

With the text cleaned up and pricing data retrieved, we then dropped any columns with missing pricing data or text \footnote{This left us with 392 of 509 S\&P constituents. Informally, there were still plenty of examples from each sector, but in future work, it would be a good idea to perform a rigorous statistical sampling to avoid overweighting any one market sector.} into an sqlite3 database for scalability and portability.

\section{Approach}
To test whether or not using NLP sentiment analysis on stock transcript data is effective or not, we decided to approach the problem using multiple existing models re-trained and tuned on our custom dataset. The original model selections were BERT\cite{BERT}, FinBERT\cite{FinBERT1}, and ULMFiT\cite{ULMFit}. After further testing, Longformer\cite{Longformer} was also selected for analysis.

\subsection{Architecture Selection}
We primarily selected transformers for this task for a variety of reasons. The attention mechanisms built into these models allow for us to pay closer attention to features in the transcripts that have more weight in the final classification. We are also able to utilize long range dependencies between the elements, which allows us to handle the large token length in our input data.

\subsection{Data Preparation}
We had to use even more deep learning concepts in our preparation of the data. We had to properly label, tokenize, and split our data in preparation for our training.

Our truth labels were calculated using total price movement of the share price in the days before and the days after using the following equation:

\begin{equation*}
price\_movement = \left(\frac{{\text{{sp}}_{t+2} - \text{{sp}}_{t-2}}}{{\text{{sp}}_{t-2}}} \right) \times 100
\end{equation*}
\begin{itemize}
    \item Where $\text {sp}_{t+2}$ is the closing share price 2 days after the report.
    \item Where $\text{sp}_{t-2}$ is the closing share price 2 days before the report.
\end{itemize}

\begin{figure}[h!]
    \centering
    \begin{subfigure}{0.5\linewidth}
        \centering
        \includegraphics[width=\linewidth]{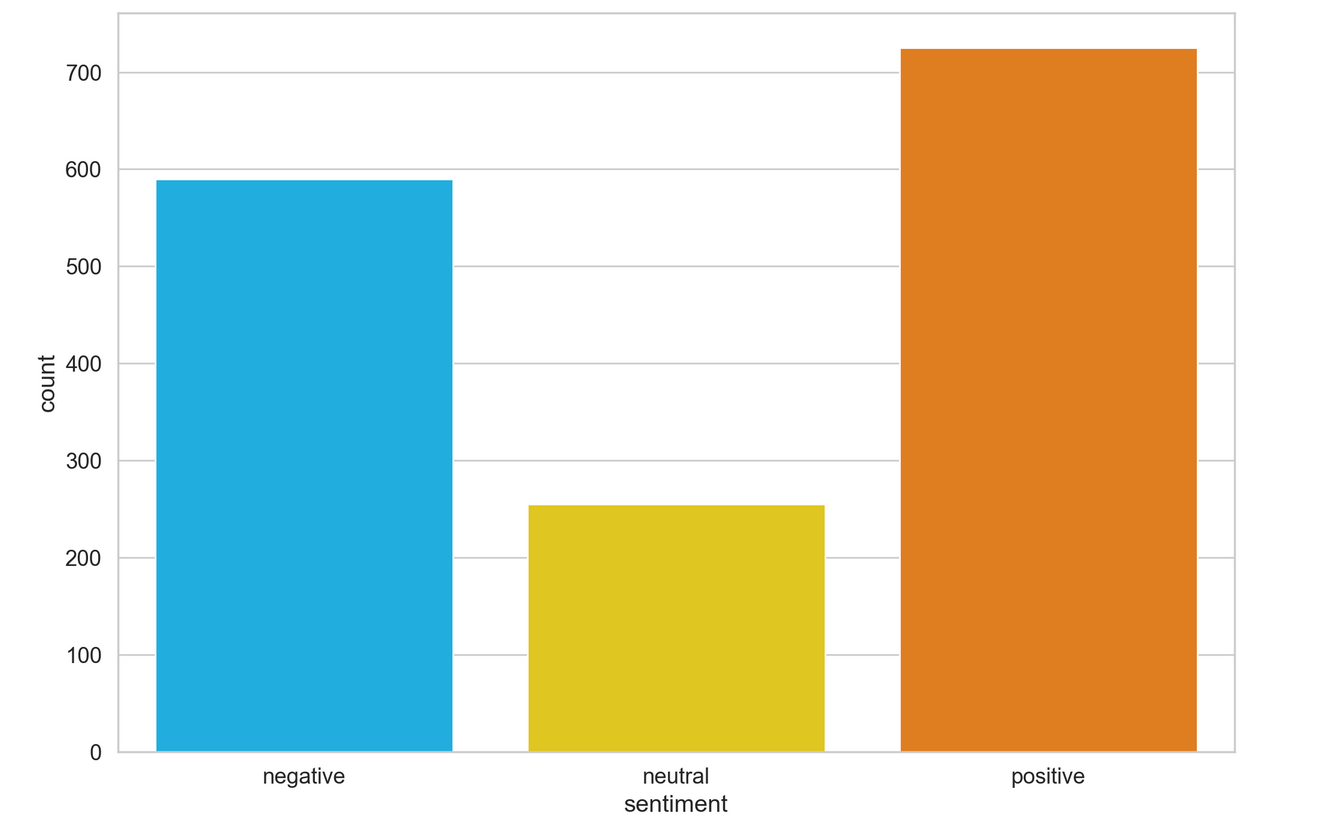}
        \caption{Truth label class balance}
        \label{fig:truth_label_class_balance}
    \end{subfigure}%
    \begin{subfigure}{0.5\linewidth}
        \centering
        \includegraphics[width=\linewidth]{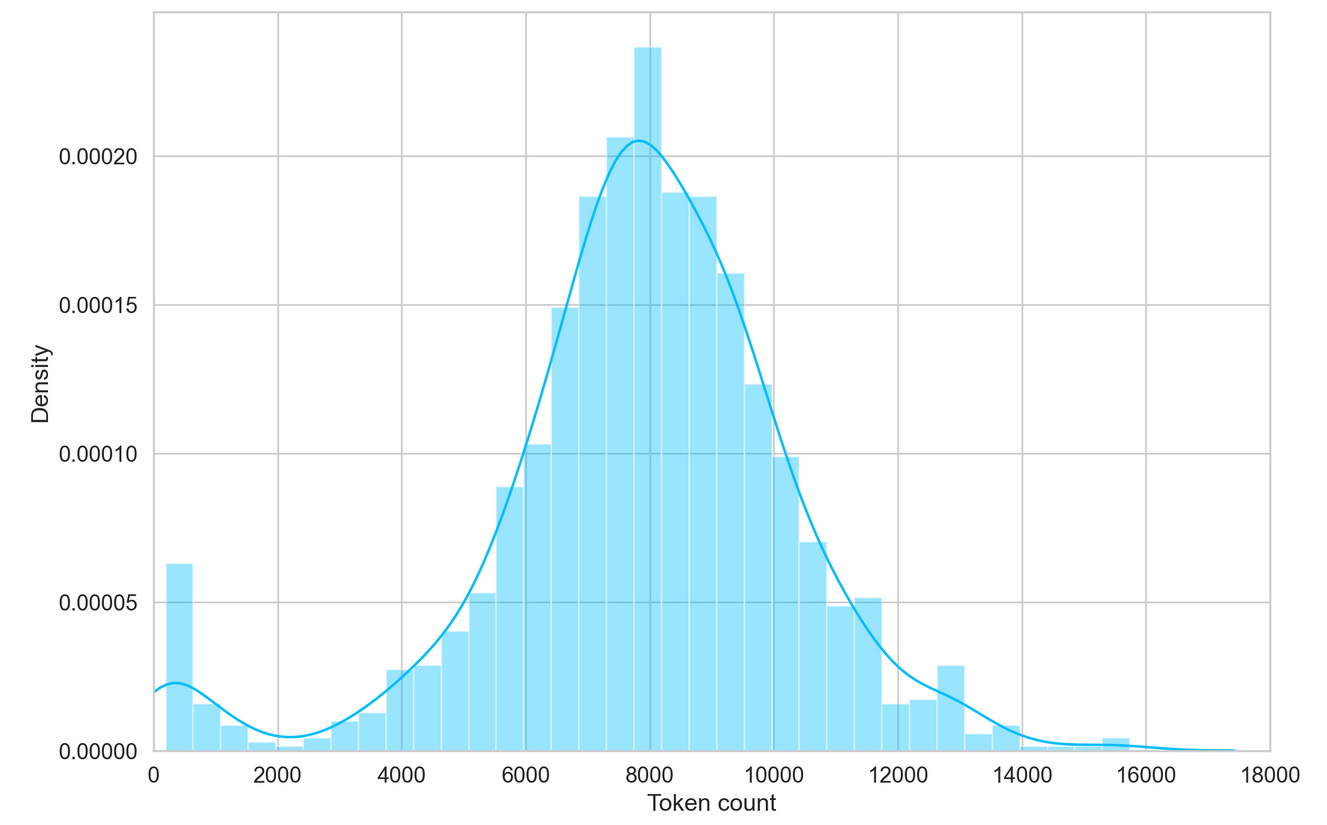}
        \caption{Token count density}
        \label{fig:token_count_density}
    \end{subfigure}
    \caption{Data Preparation Plots}
    \label{fig:data_prep_plots}
\end{figure}

We then checked if the price movement was less than a negative threshold, between a negative and a positive threshold, or greater than a positive threshold. This gave us our truth labels (negative = 0, neutral = 1, and positive = 2, respectively). A negative threshold of -3 and a positive threshold of 3 separated our data into fairly good distributions of classes(\ref{fig:truth_label_class_balance}).

After assigning truth labels to all transcripts for our data points, we then split the data into train, validation, and test sets. When we encoded the data and looked at the distribution of token lengths for our dataset(\ref{fig:token_count_density}), we were hit with our first big problem.

The token token count went as high as 16000 tokens. For many of our model's the max sequence length intended for ingestion was 512.

To handle the problem with the large amount of tokens in each data input, we looked at a variety of solutions. The first solution was to truncate the data to the maximum length that the models could handle. Although this was enough for some of our models, we did lose a lot of potentially important features of the data.

Another solution some models took was to separate each transcript into chunks of the maximum token length for that model. This method was very computationally expensive and required long hours of training on very expensive GPUs. This method led us to also test out the Longformer model, which had built-in methods for handling large amounts of text.

\subsection{Anticipated Goals and Challenges}
By re-training these models on our custom stock transcript dataset, we believed that we would be able to provide beneficial insights on how effective these NLP models are at handling large amounts of financial data and making stock market predictions based on sentiment analysis. Although we anticipated issues with large datasets, we did not anticipate how difficult it would be to get a helpful sentiment analysis from the type of data we were looking at. The effects of the "sugar coating" presented by the companies proved to be a difficult challenge to overcome.

\section{Experiments and Results}
\subsection{FinBERT Model}
\subsubsection{Overview and Experimental Procedure}
The FinBERT model is a variant of the BERT model pre-trained on a large corpus of financial texts. This pre-training ensures that the model is equipped with domain-specific knowledge that can lead to better performance in capturing financial sentiment expressions\cite{FinBERT1}. The data is represented through tokenization, where each word in the text is mapped to a unique token. These tokens represent the input fed into the model. FinBERT consists of multiple layers of self-attention mechanisms and feedforward neural networks that efficiently capture the bidirectional contextual relationship between the tokens during the forward and backward passes. The output is the probability distribution for the possible sentiment classes (0: negative, 1: neutral, 2: positive).
The cross-entropy loss function was chosen as it is well-suited for multi-class classification problems like sentiment analysis. The initial optimizer used was the AdamW due to its effectiveness in optimizing deep learning models with large parameter spaces.
We used the PyTorch deep learning framework and sourced the pre-trained FinBERT model from the Hugging Face repository.

In this experiment\cite{final_project_source_code}, since FinBERT is a large model, thus computationally expensive, we employed various strategies such as reducing batch size, using more powerful GPUs (e.g., Nvidia A100), and optimizing memory usage. We mitigated the problem of the length limit by tokenizing the data, splitting it into chunks, assigning IDs to chunks to regroup into transcripts after evaluation, and modifying batching techniques. We explored different sequence length limits (128, 256, 512).

After dividing our transcripts into training, validation, and testing sets, we selected a baseline where the (not fine-tuned) pre-trained FinBERT is evaluated on our testing set.

\begin{figure}[h!]
    \centering
    \begin{subfigure}{0.5\linewidth}
        \centering
        \includegraphics[width=\linewidth]{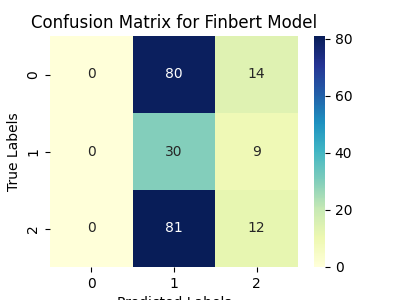}
        \caption{Confusion matrix}
        \label{fig:conf_baseline_finbert}
    \end{subfigure}%
    \begin{subfigure}{0.5\linewidth}
        \centering
        \includegraphics[width=\linewidth]{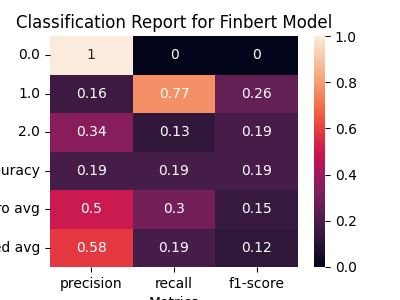}
        \caption{Classification report}
        \label{fig:report_baseline_finbert}
    \end{subfigure}
    \caption{Performance Metrics for baseline FinBERT}
    \label{fig:baseline-finbert-reports}
\end{figure}
\begin{figure}[h!]
    \centering
    \begin{subfigure}{0.4\linewidth}
        \centering
        \includegraphics[width=\linewidth]{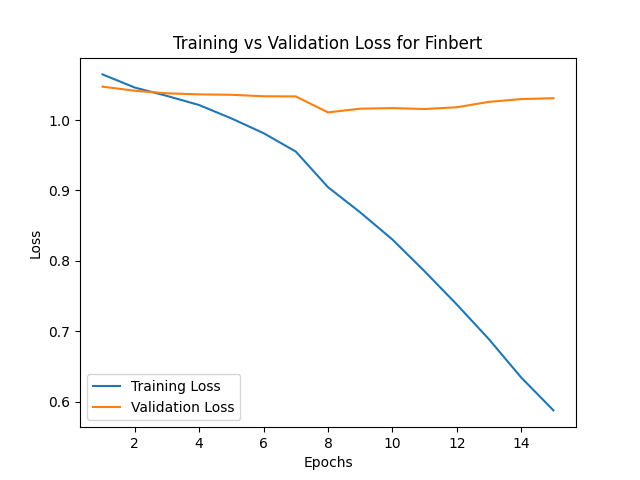}
        \caption{Training vs. \\ Validation Loss}
        \label{fig:losses_finbert}
    \end{subfigure}%
    \begin{subfigure}{0.4\linewidth}
        \centering
        \includegraphics[width=\linewidth]{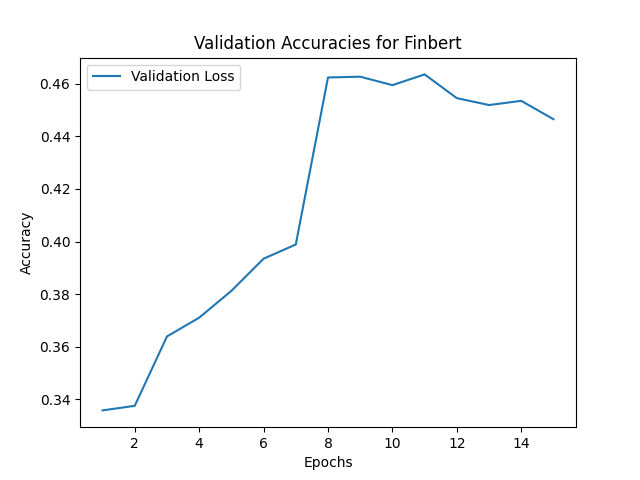}
        \caption{Validation \\Accuracy}
        \label{fig:accurs_finbert}
    \end{subfigure}
    \caption{Training Metrics for fine-tuned FinBERT}
    \label{fig:finbert-losses-accurs}
\end{figure}

We then proceeded to fine-tune the FinBERT model. As the model initially struggled to learn, we experimented with text cleanup techniques, modifying loss functions (hinge loss), different number of classes, and focusing on different sections of the transcripts. After extensive fine-tuning, we ended up with the following optimized configurations:\\
\textit{Model}: “yiyanghkust/finbert-pretrain”\cite{FinBERT2}\\
\textit{Input}: Q$\&$A sections of the transcripts to reduce noise.\\
\textit{Loss Function}: Use of Weighted Cross-Entropy (with heavy penalties for misclassifying negative and neutral labels). After 6 epochs, we switched to regular Cross-Entropy loss.\\
\textit{Optimizer}: Removed the scheduler and switched to RMSProp to increase stability during training and mitigate sparse gradients in large vocabulary.\\
\textit{Hyperparameters}: \\
-	Batch Size = 8 (representing number of transcripts and not chunks)\\
-	Number of Epochs = 15:
Less than 5 epochs would cause all transcripts to be labeled as positive and over 15 epochs would show pronounced over-fitting(\ref{fig:losses_finbert}).\\
-	Learning Rate = 0.00001 to help preserve the pre-trained knowledge\\ 
We aggregated the labels by selecting the most occurring label in each chunked transcript and obtained the following results.

\begin{figure}[h!]
    \centering
    \begin{subfigure}{0.5\linewidth}
        \centering
        \includegraphics[width=\linewidth]{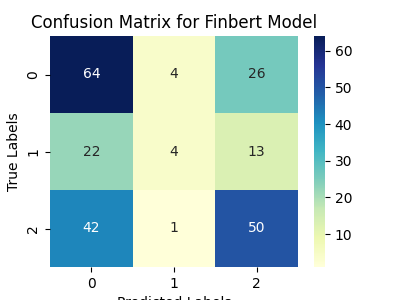}
        \caption{Confusion matrix}
        \label{fig:class_finbert}
    \end{subfigure}%
    \begin{subfigure}{0.5\linewidth}
        \centering
        \includegraphics[width=\linewidth]{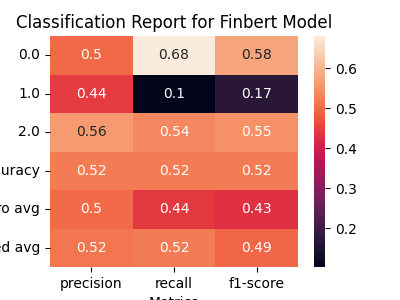}
        \caption{Classification report}
        \label{fig:conf_finbert}
    \end{subfigure}
    \caption{Performance Metrics for fine-tuned FinBERT}
    \label{fig:finbert-reports}
\end{figure}

\begin{figure}[h]
    \centering
    \includegraphics[width=0.4\linewidth]{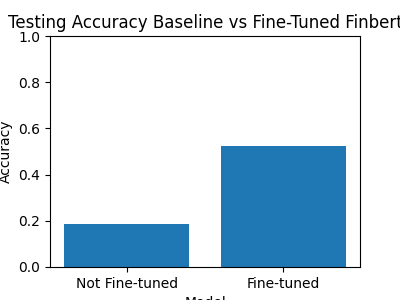}
    \caption{Testing Accuracy Baseline vs Fine-Tuned FinBERT}
    \label{fig:test_accur_finbert}
\end{figure}
\subsubsection{Analysis}
The results from the baseline model reveal a struggle to effectively capture the nuanced language patterns inherent to our earnings transcripts texts probably due to a lack of task-specific knowledge. It achieved an accuracy of 18.58$\%$, with a weighted average F1-score of 12.22$\%$(\ref{fig:report_baseline_finbert}). The low accuracy and F1-score reflect a generalization gap between the original pre-training data and the target task. Upon further analysis, all transcripts were labeled as neutral and positive, and the model failed to pick up a negative tone(\ref{fig:conf_baseline_finbert}). 

After fine-tuning, the FinBERT model was able to adapt its learned representations to better suit the specific characteristics of our earnings call transcripts task. The accuracy dramatically increased to 52.21$\%$(\ref{fig:accurs_finbert}). The confusion matrix and classification report indicate a more balanced distribution of correct predictions across sentiment classes, with improvements observed in precision (~.52), recall (~.52), and F1-scores (~.49) for each class (including the ability to recognize negative sentiments)(\ref{fig:finbert-reports}). 

Through this transfer learning method, the FinBERT model was able to focus on the complex relationship between sentiment and financial terminology in the context of earning calls and updated its parameters with task-specific data while still leveraging its prior knowledge of the financial domain.

\subsection{BERT Model}\label{subsec:bert_section}
\subsubsection{Overview and Experimental Procedure}
The BERT (Bidirectional Encoder Representations from Transformers) model is an NLP model pre-trained by Google\cite{BERT_model} for handling text data. It is a well known transformer, and can easily be fine-tuned for new data with the addition of one output layer. This model was selected due to its transformer architecture and effectiveness in tasks such as sentiment analysis. Some example code from a base implementation of BERT\cite{base_BERT_implementation} was used as starting source code. Our source code can be found on GitHub\cite{final_project_source_code}.

Our objective was to leverage transfer learning techniques on the base BERT model to extract sentiment analysis from ingested transcript data. To do this, we built a sentiment classifier on top of the basic BERT model. We used a dropout layer for regularization and a fully-connected output layer. We utilized an optimizer that was a version of the Adam algorithm with added weight decay fixing. During the training, we utilized cross-entropy loss as our loss function. All of the code for the modified BERT was implemented using PyTorch as the Deep Learning framework. For the BERT model, the data was split into an 80\% training set, 10\% validation set, and 10\% testing set.

The main limitation for BERT is the max sequence length of 512. Due to the large token length of the data, we had to either chunk it into manageable segments or truncate it. The amount of GPU memory utilized to chunk the data proved to be the main limitation. It turned out for this model, that truncating and additional parsing seemed to deliver the best results.

\subsubsection{Analysis}
The data was limited to the first 512 tokens of the Q\&A section of the transcripts. After many training runs, the following hyper parameters were selected:
Batch Size = 32, 
Epochs = 10, 
Learning Rate = 0.00001.

The higher batch size is probably the reason this method performed better for BERT then some of the other models. When separating each transcript into manageable segments with a batch size of 32, even more powerful GPUs ran out of memory. By truncating our inputs, we were able to use a higher batch size and assist with the intense over fitting by potentially acting as a form of regularization (averaging gradients over more examples).

\begin{figure}[h!]
    \centering
    \begin{subfigure}{0.4\linewidth}
        \centering
        \includegraphics[width=\linewidth]{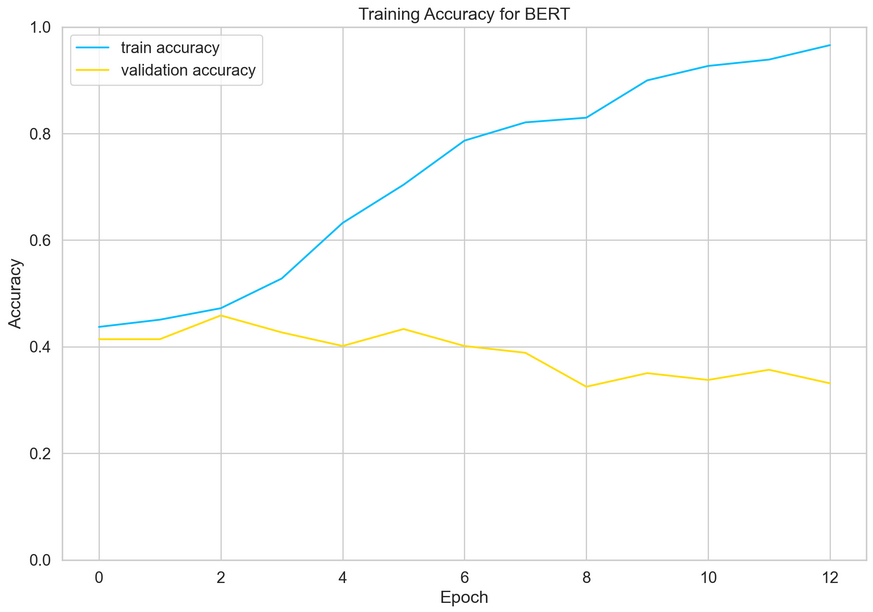}
        \caption{Training vs. \\ Validation Accuracy}
        \label{fig:bert_training_acc}
    \end{subfigure}%
    \begin{subfigure}{0.4\linewidth}
        \centering
        \includegraphics[width=\linewidth]{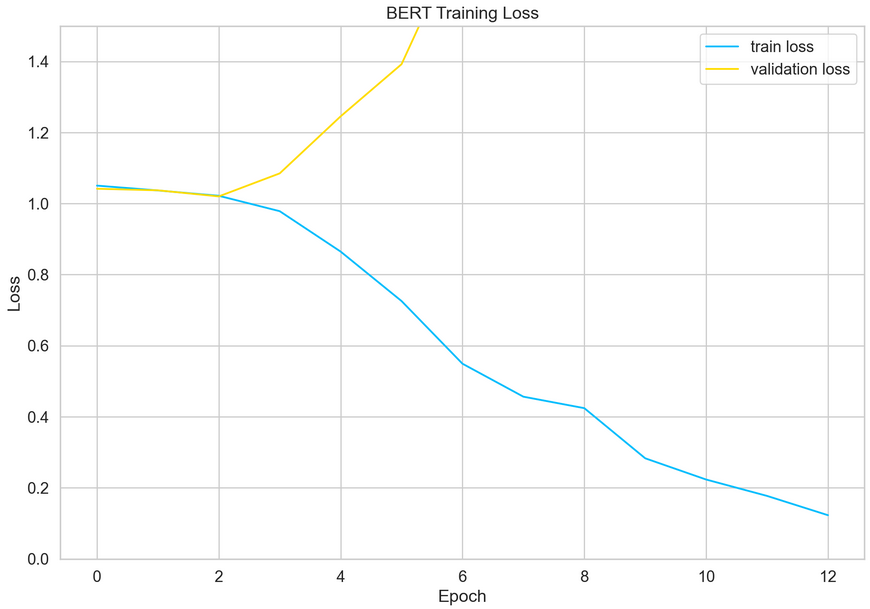}
        \caption{Training vs. \\ Validation Loss}
        \label{fig:bert_training_loss}
    \end{subfigure}
    \caption{Training Metrics for fine-tuned BERT}
    \label{fig:bert_trianing_metrics}
\end{figure}

When lowering the Epochs to 5, the data tended to have a good score, but would only output positive labels when predicting. This was likely due to the "sugar coating" problem discussed earlier. The model likely needed more epochs to adjust for this uniqueness in the data(\ref{fig:bert_training_acc}). Changing the learning rate did not change the results too much in the model though. Using these hyper parameters, the following results were extracted.

\begin{figure}[h!]
    \centering
    \begin{subfigure}{0.47\linewidth}
        \centering
        \includegraphics[width=\linewidth]{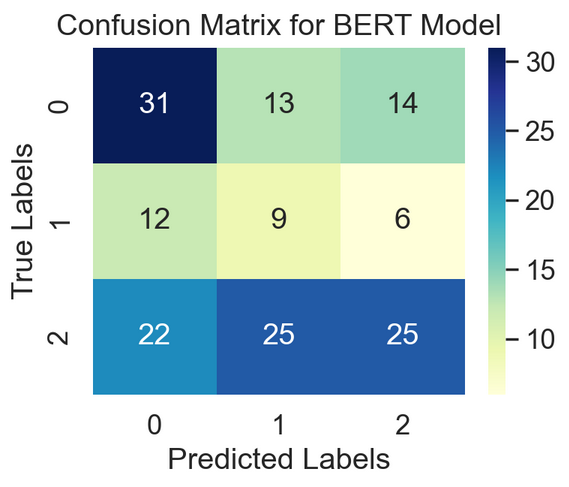}
        \caption{Confusion matrix}
        \label{fig:bert_confusion_matrix}
    \end{subfigure}%
    \begin{subfigure}{0.60\linewidth}
        \centering
        \includegraphics[width=\linewidth]{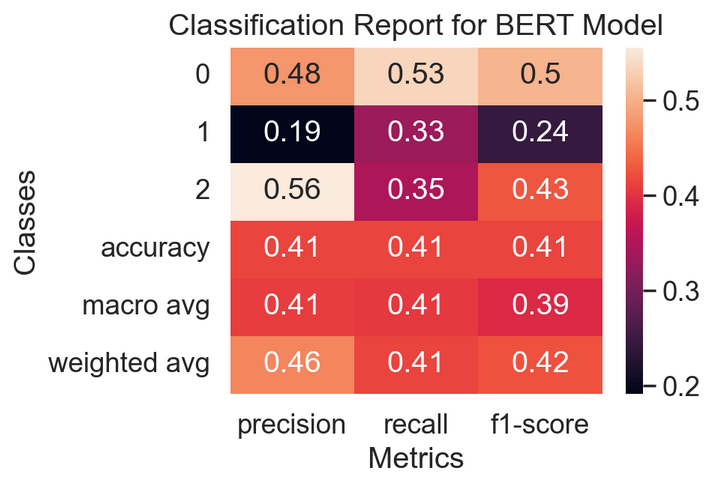}
        \caption{Classification report}
        \label{fig:bert_class_report}
    \end{subfigure}
    \caption{Performance Metrics for fine-tuned BERT}
    \label{fig:bert-metrics}
\end{figure}

The model did not seem to be very capable of predicting price movement in stocks based off of these heavily-biased transcripts. The length of the data, paired with the sugar coating, provided very steep challenges to overcome. Though the model did end up with a decent distribution of positive, negative, and neutral results(\ref{fig:bert_confusion_matrix}), it struggled with the correlation of sentiment to price movement.

The model ended up with a test accuracy of roughly 41\% and a weighted average F1 score of around 42\%(\ref{fig:bert_class_report}). These are fairly low scores, but considering the challenges in this problem, and comparing it to previous runs before the final fine-tuned model was trained, BERT did very well. It certainly discovered a slight relationship between equity earnings call transcripts and the future performance of the company, but it failed to fully extract the most relevant pieces of data from all the biased data mixed in.

\subsection{ULMFiT Model}
\subsubsection{Overview and Experimental Procedure}
One of the primary challenges in deep learning is the substantial data requirements for training models. We encountered this challenge with earnings transcripts data, which, while extensive in textual format, was constrained by a limited number of stocks. To address this limitation, we employed the concept of transfer learning and chose ULMFiT (Universal Language Fine-tuning for Text Classification) for its efficacy in transfer learning for text classification. We believed that by utilizing a pre-trained ULMFiT model and training it on our financial transcript data, we could capitalize on its ability to capture semantic representations and linguistic nuances, thus potentially improving classification performance.\cite{howard2018ulmfit} 

We adopted a two-tier framework for text classification. Initially, a pre-trained ULMFiT language model was employed, trained on a large corpus. This model was pre-trained on a large dataset of text, such as Wikipedia or Yelp, aiding transfer learning. Using the model's understanding of natural language, we trained sentiment analysis on financial transcripts. Finally, the vocabulary learned by the language model and its encoder were utilized by the classification model, providing access to semantic representations\cite{SentimentAnalysis11}. 

The model structure consists of SequentialRNN architecture, specifically an LSTM-based recurrent neural network. The model structure includes multiple LSTM layers followed by dropout layers and batch normalization. The ouput layer consists of a linear layer with a ReLU activation function, followed by another layer for classification. The total number of trainable parameters when using the whole transcript text data for each model is 62,650. The output is a softmax layer output that was classified in the labels indicating price movement categories (0-negative, 1-positive, 2-neutral). The loss function and Optimizer being used is CrossEntropyLoss and Adam optimizer. The best hyperparameters were chosen for learning rate (0.00173), batch size (64), epochs (30), dropout rate (0.3).

\subsubsection{Analysis} 
\begin{figure}[h!]
    \centering
    \begin{subfigure}{0.45\linewidth}
        \centering
        \includegraphics[width=\linewidth]{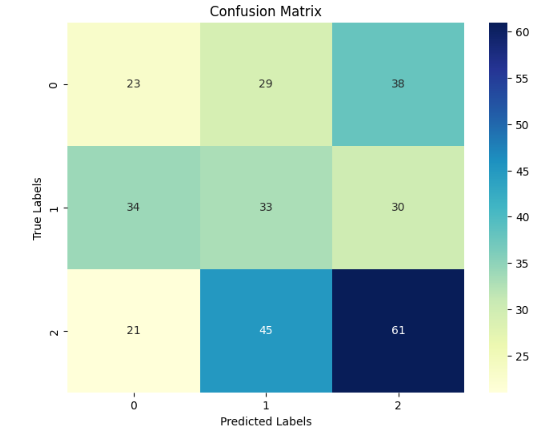}
        \caption{Confusion matrix}
        \label{fig:confusion-matrix}
    \end{subfigure}%
    \begin{subfigure}{0.45\linewidth}
        \centering
        \includegraphics[width=\linewidth]{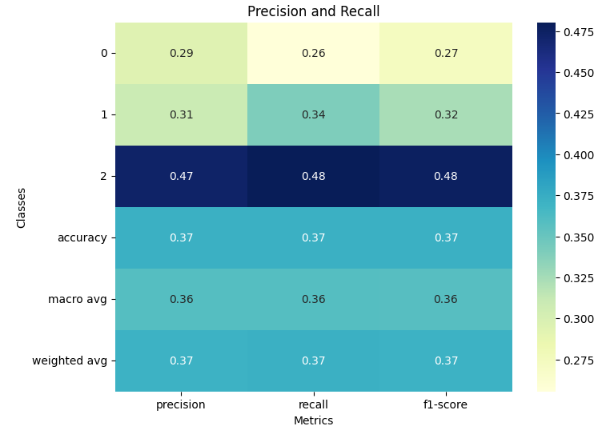}
        \caption{Classification report}
        \label{fig:precision-recall}
    \end{subfigure}
    \caption{Performance Metrics for Full-Transcript ULMFit}
    \label{fig:ulmfit-metrics}
\end{figure}

\begin{figure}[h!]
    \centering
    \begin{subfigure}{0.45\linewidth}
        \centering
        \includegraphics[width=\linewidth]{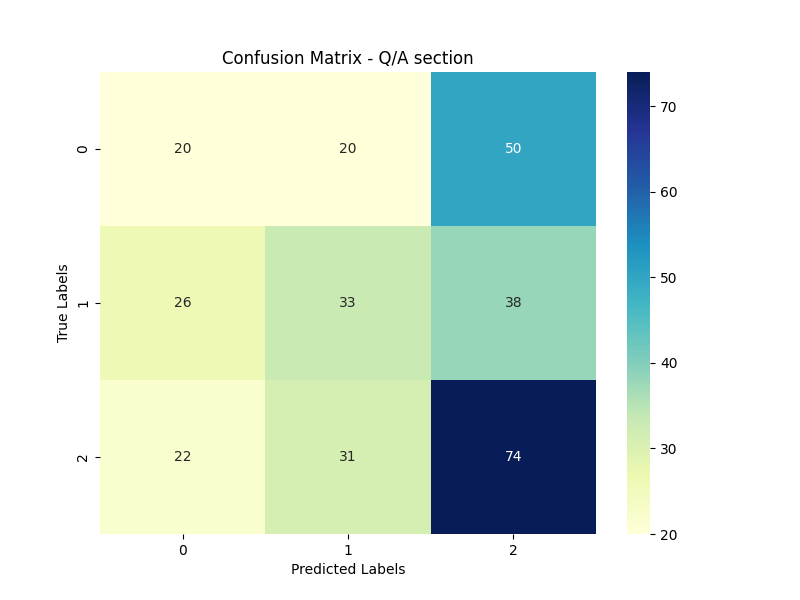}
        \caption{Confusion matrix}
        \label{fig:confusion-matrix-qa}
    \end{subfigure}%
    \begin{subfigure}{0.45\linewidth}
        \centering
        \includegraphics[width=\linewidth]{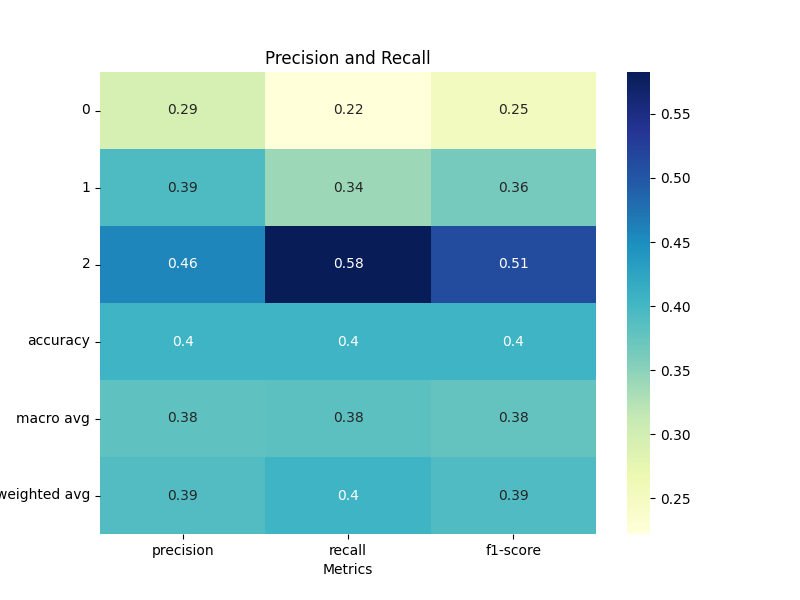}
        \caption{Classification report}
        \label{fig:precision-recall-qa}
    \end{subfigure}
    \caption{Performance Metrics for Parsed Q\&A Transcript ULMFit}
    \label{fig:ulmfit-metrics-qa}
\end{figure}

The Overall observed accuracy was 37\%(\ref{fig:ulmfit-metrics}) when using the entire raw transcript data. The precision and recall were around 37\%. In order to improve the results and provide a better context driven text, we used the Q\&A section within the transcript text data to train our models. The results improved as the over fitting reduced in relative to using the overall transcript data. The observed accuracy was around 40\% with precision, F1 around 39\% and recall around 40\%(\ref{fig:ulmfit-metrics-qa}). 

However, computational challenges and long training times were encountered due to the model complexity and transcript data size. One method employed to address GPU memory limitations and improve training efficiency was data segmentation into chunks. 

\begin{figure}[h!]
    \centering
    \includegraphics[width=0.7\linewidth]{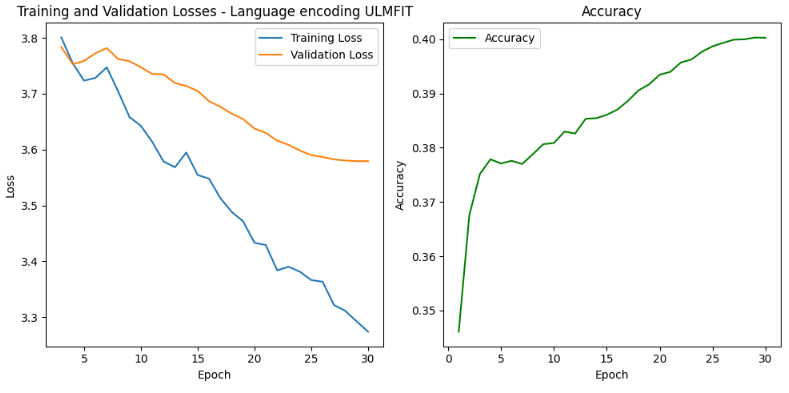}
    \caption{Training Metrics for Fine-tuned ULMFit}
    \label{fig:ulm_train_loss_acc}
\end{figure}

Overall, the results showed discrepancy between training and validation losses(\ref{fig:ulm_train_loss_acc}) which suggests potential over fitting issues for the classification model especially when training on the entire transcript. When focused on Q\&A part, the validation loss and accuracy improved. However, it was still lower than optimal. This suggests that ULMFiT was still unable to fully capture and represent the nuances present in text data. The source code can be found on Github.\cite{final_project_source_code}


\subsection{Longformer Model}
The Longformer model or Long-Document Transformer is a pre-trained NLP model trained by AllenAI\cite{longformer_model}. This model was selected due to its transformer architecture and ability to ingest a text data with a max token length of 4096 efficiently.

Like the others, it was not very capable of predicting price movement from the heavily-biased transcripts. Even though more data was able to be processed, the sugar coating issue still persisted in causing this model to struggle. See the Appendix(\ref{longformer_appendix}) for additional details on this model.

\section{Conclusion}
Overall, our results showed BERT-base model performed better than ULMFiT. The bidirectional context understanding in BERT, facilitated by its deep transformer architecture with multiple layers and attention mechanisms, enabled it to capture complex linguistic patterns and relationships. In contrast, ULMFiT uses a simple architecture based on RNN’s (LSTM) that did not enable it to learn complex patterns relative to BERT. Similar to BERT, FinBERT leveraged BERT’s complex architecture while incorporating domain-specific modifications and is able to perform the best for our text classification and sentiment analysis on earnings call transcript data.

One of the key challenges in dealing with the earnings call transcript data, the language used often employs sugar-coated rhetoric making it challenging for any sentiment analysis model to accurately discern negative truths. Phrases like \textbf{"By addressing costs and margins, we aim to drive future success and leverage our innovative offerings to create value for our customers"} are presented in a way that emphasizes positivity and confidence in future outcomes leading the model to inaccurately classify the negative sentiment, as it may be masked by optimistic rhetoric. \\

\section{Future Work}

For this project, we retreived the earnings reports for (approximately) FY2023. Since FY23 was a bullish year for large-cap equities, It's possible that the participants in these earnings calls were broadly more optimistic than other years. To test this hypothesis, it would be a good idea to take samples from a mediocre year (FY2016) and an \textit{awful} year (FY2009). By virtue of being included in the S\&P 500 Index, the dataset is comprised entirely large-cap U.S. equities. International, small- and mid-cap equities could be included in future work.

The prevalent theme in this paper is the tendency to sugarcoat bad news, which throws off a general-purpose sentiment analysis. By building a very large data set of most or all earnings calls from publicly traded companies, it might be possible to build a BERT-like model \textit{exclusively} from these transcripts, rather than being pre-trained on broader corpora. We believe this could better capture the 'corporatese' of a typical earnings call.

\newpage

\newpage
\section{Appendix}
\subsection{Longformer Model}
\label{longformer_appendix}
\subsubsection{Overview and Experimental Procedure}
The Longformer model or Long-Document Transformer is a pre-trained NLP model trained by AllenAI\cite{longformer_model}. This model was selected due to its transformer architecture and ability to ingest a text data with a max token length of 4096. Our source code can be found on GitHub\cite{final_project_source_code}.

Training Longformer followed the same setup as BERT(\ref{subsec:bert_section}) Only modifications were made to the max length and the hyper parameters.

Since the token length was much higher, we decided to truncate the data at the maximum length. This was mostly due to the higher memory requirements of Longformer and the fact that the model already divides the input sequence into windows to handle them properly.

\subsubsection{Fine-tuned Longformer Analysis}
After many training runs, the following hyper parameters were selected:
Batch Size = 4, 
Epochs = 10, 
Learning Rate = 0.00001.

Due to the computationally expensive nature of Longformer, the batch size had to be very low. Similar to some of the other models, when lowering the Epochs to 5, the model performed well qualitatively, but in reality it would only output positive classifications. These hyper parameters gave us the following results.

\begin{figure}[h!]
    \centering
    \begin{subfigure}{0.47\linewidth}
        \centering
        \includegraphics[width=\linewidth]{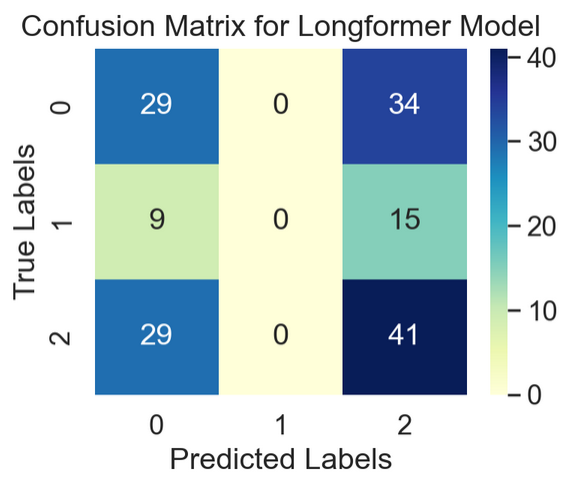}
        \caption{Confusion matrix}
        \label{fig:longformer_confusion_matrix}
    \end{subfigure}%
    \begin{subfigure}{0.6\linewidth}
        \centering
        \includegraphics[width=\linewidth]{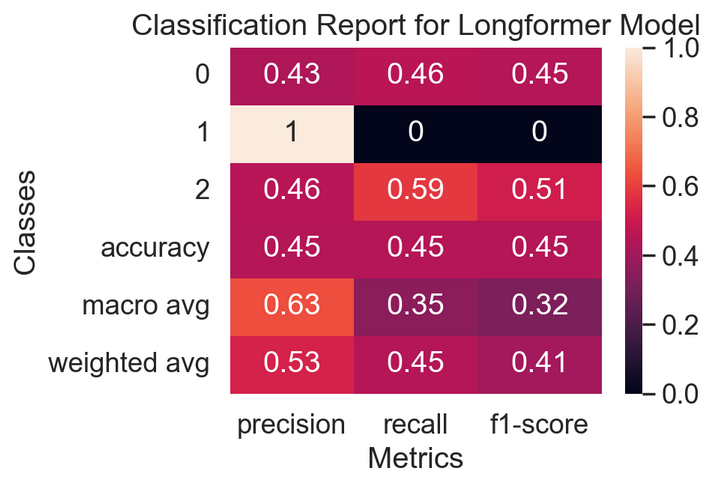}
        \caption{Classification report}
        \label{fig:longformer_class_report}
    \end{subfigure}
    \caption{Performance Metrics for fine-tuned Longformer}
    \label{fig:longformer-metrics}
\end{figure}

\begin{figure}[h!]
    \centering
    \begin{subfigure}{0.4\linewidth}
        \centering
        \includegraphics[width=\linewidth]{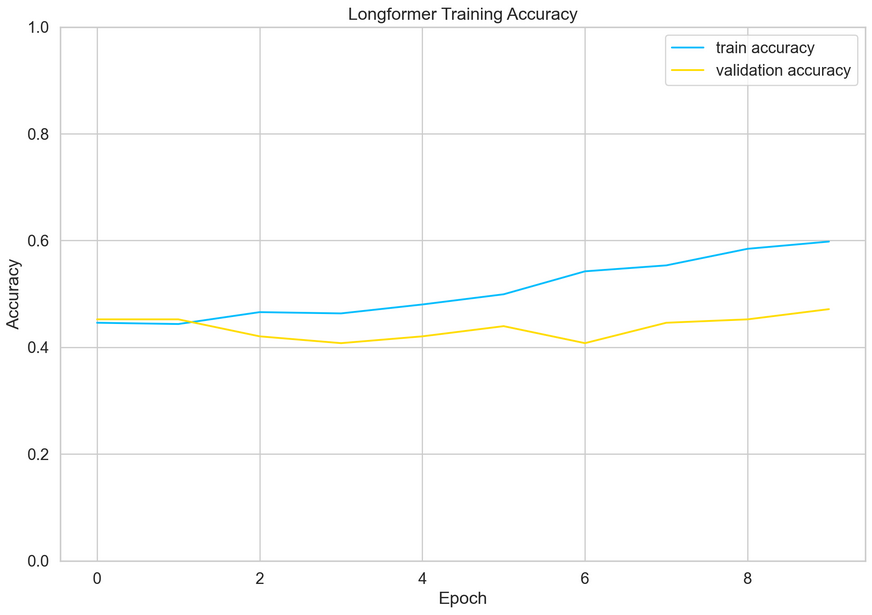}
        \caption{Training vs. \\ Validation Accuracy}
        \label{fig:longformer_training_acc}
    \end{subfigure}%
    \begin{subfigure}{0.4\linewidth}
        \centering
        \includegraphics[width=\linewidth]{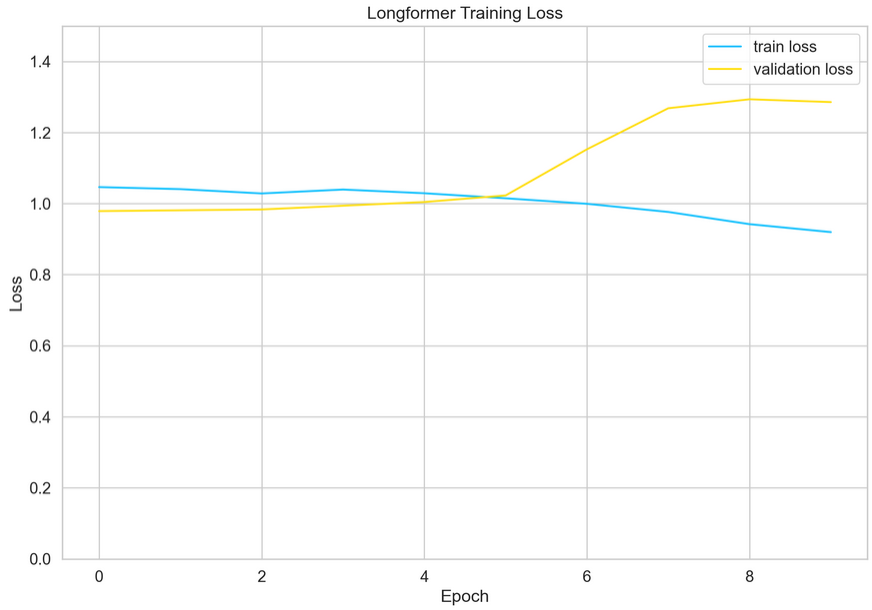}
        \caption{Training vs. \\ Validation Loss}
        \label{fig:longformer_training_loss}
    \end{subfigure}
    \caption{Training Metrics for fine-tuned Longformer}
    \label{fig:longformer_trianing_metrics}
\end{figure}

Though Longformer was capable of handling larger segments of data fairly efficiently, it did not seem to improve upon the other models. Like the others, it was not very capable of predicting price movement from the heavily-biased transcripts. Even though more data was able to be processed, the sugar coating issue still persisted in causing this model to struggle.

The model ended up with a test accuracy of roughly 45\% and a weighted average F1 score of around 41\%. Though it seemed to improve over some of the other models as far as weighted average, the macro average metrics dropped considerably due to the lack of neutral classifications.

\newpage
 
\newpage

{\small
\bibliographystyle{ieee_fullname}
\bibliography{egbib}

}

\end{document}